# Playing for Benchmarks

Stephan R. Richter  Zeeshan Hayder  Vladlen Koltun
TU Darmstadt       ANU             Intel Labs

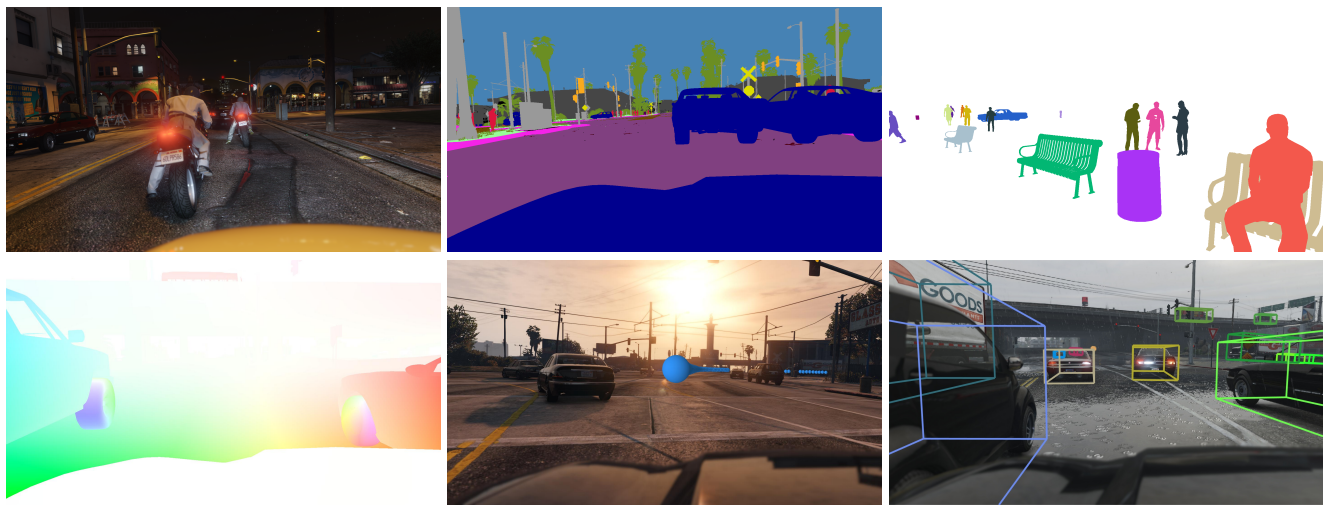

Figure 1. Data for several tasks in our benchmark suite. Clockwise from top left: input video frame, semantic segmentation, semantic instance segmentation, 3D scene layout, visual odometry, optical flow. Each task is presented on a different image.

## Abstract

*We present a benchmark suite for visual perception. The benchmark is based on more than 250K high-resolution video frames, all annotated with ground-truth data for both low-level and high-level vision tasks, including optical flow, semantic instance segmentation, object detection and tracking, object-level 3D scene layout, and visual odometry. Ground-truth data for all tasks is available for every frame. The data was collected while driving, riding, and walking a total of 184 kilometers in diverse ambient conditions in a realistic virtual world. To create the benchmark, we have developed a new approach to collecting ground-truth data from simulated worlds without access to their source code or content. We conduct statistical analyses that show that the composition of the scenes in the benchmark closely matches the composition of corresponding physical environments. The realism of the collected data is further validated via perceptual experiments. We analyze the performance of state-of-the-art methods for multiple tasks, providing reference baselines and highlighting challenges for future research. The supplementary video can be viewed at* <https://youtu.be/T9OybWv923Y>

## 1. Introduction

Visual perception is believed to be a compound process that involves recognizing objects, estimating their three-dimensional layout in the environment, tracking their motion and the observer's own motion through time, and integrating this information into a predictive model that supports planning and action [48]. Advanced biological vision systems have a number of salient characteristics. The component processes of visual perception (including motion perception, shape perception, and recognition) operate in tandem and support each other. Visual perception integrates information over time. And it is robust to transformations of visual appearance due to environmental conditions, such as time of day and weather.

These characteristics are recognized by the computer vision and robotics communities. Researchers have argued that diverse computer vision tasks should be tackled in concert [39] and developed models that could support this [6, 33, 43]. The importance of video and motion cues is widely recognized [1, 18, 30, 49]. And robustness to environmental conditions is a long-standing challenge in the field [38, 46].

We present a benchmark suite that is guided by these considerations. The input modality is high-resolution video. Comprehensive and accurate ground truth is provided for low-level tasks such as visual odometry and optical flow as well as higher-level tasks such as semantic instance segmentation, object detection and tracking, and 3D scene layout, all on the same data. Ground truth for all tasks is available for every video frame, with pixel-level segmentations and subpixel-accurate correspondences. The data was collected in diverse environmental conditions, including night, rain, and snow. This combination of characteristics in a single benchmark aims to support the development of broad-competence visual perception systems that construct and maintain comprehensive models of their environments.

Collecting a large-scale dataset with all of these properties in the physical world would have been impossible with known techniques. To create the benchmark, we have used an open-world computer game with realistic content and appearance. The game world simulates a living city and its surroundings. The rendering engine incorporates comprehensive modeling of image formation. The training, validation, and test sets in the dataset comprise continuous video sequences with 254,064 fully annotated frames, collected while driving, riding, and walking a total of 184 kilometers in different environmental conditions.

To create the benchmark, we have developed a new methodology for collecting data from simulated worlds without access to their source code or content. Our approach integrates dynamic software updating, bytecode rewriting, and bytecode analysis, going significantly beyond prior work to allow video-rate collection of ground truth for all tasks, including subpixel-accurate dense correspondences and instance-level 3D layouts.

We conduct extensive experiments that evaluate the performance of state-of-the-art models for semantic segmentation, semantic instance segmentation, visual odometry, and optical flow estimation on the presented benchmark. The results indicate that the benchmark is challenging and creates new opportunities for progress. Detailed performance analyses point to promising directions for future work.

## 2. Background

Progress in computer vision has been driven by the systematic application of the common task framework [13]. The Pascal VOC benchmark supported the development of object detection techniques that broadly influenced the field [16]. The ImageNet benchmark was instrumental in the development of deep networks for visual recognition [56]. The Microsoft COCO benchmark provides data for object detection and semantic instance segmentation [35]. The SUN and Places datasets support research on scene recognition [62, 66]. The Middlebury benchmarks for stereo, multi-view stereo, and optical flow played a pivotal role in the development of low-level vision algorithms [4, 57, 58, 59].

The KITTI benchmark suite provides ground truth for visual odometry, stereo reconstruction, optical flow, scene flow, and object detection and tracking [19]. It is an important precursor to our work because it provides video input and evaluates both low-level and high-level vision tasks on the same data. It also highlights the limitations of conventional ground-truth acquisition techniques [21]. For example, optical flow ground truth is sparse and is based on fitting approximate CAD models to moving objects [42], object detection and segmentation data is available for only a small number of frames and a small set of classes [63], and the dataset as a whole represents the appearance of a single town in fair weather.

The Cityscapes benchmark evaluates semantic segmentation and semantic instance segmentation models on images acquired while driving around 50 cities in Europe [10]. The quality of the annotations is very high and the dataset has become the default benchmark for semantic segmentation. It also highlights the challenges of annotating real-world images: only a sparse set of 5,000 frames is annotated at high accuracy, and average annotation time was more than 90 minutes per frame. Data for tasks other than semantic segmentation and semantic instance segmentation is not provided, and data was only acquired during the day and in fair weather.

The importance of robustness to different environmental conditions is recognized as a key challenge in the autonomous driving community. The HCI benchmark suite offers recordings of the same street block on six days distributed over three seasons [34]. The Oxford RobotCar dataset provides more than 100 recordings of a route through central Oxford collected over a year, including recordings made at night, under heavy rain, and in the presence of snow [38]. These datasets aim to address an important gap, but are themselves limited in ways that are indicative of the challenges of ground-truth data collection in the physical world. The HCI benchmark lacks ground truth for moving objects, does not address tasks beyond stereo and optical flow, and is limited to a single 300-meter street section. The Oxford RobotCar dataset only provides GPS, IMU, and LIDAR traces with minimal post-processing and does not contain the ground-truth data that would be necessary to benchmark most tasks considered in this paper.

Our benchmark combines and extends some of the most compelling characteristics of prior benchmarks and datasets for visual perception in urban environments: a broad range of tasks spanning low-level and high-level vision evaluated on the same data (KITTI), highly accurate instance-level semantic annotations (Cityscapes), and diverse environmental conditions (Oxford). We integrate these characteristics in a single large-scale dataset for the first time, and go beyond by providing temporally consistent object-level semantic ground truth and 3D scene layouts at video rate, as well as subpixel-accurate dense correspondences. To achieve this, we use three-dimensional virtual worlds.

Simulated worlds are commonly used to benchmark optical flow algorithms [5, 4, 8] and visual odometry systems [22, 24, 64]. Virtual worlds have been used to evaluate the robustness of feature descriptors [31], test visual surveillance systems [60], evaluate multi-object tracking models [17], and benchmark UAV target tracking [44]. Virtual environments have also been used to collect training data for pedestrian detection [40], stereo reconstruction and optical flow estimation [41], and semantic segmentation [23, 55].

A key challenge in scaling up this approach to comprehensive evaluation of broad-competence visual perception systems is populating virtual worlds with content: acquiring and laying out geometric models, applying surface materials, configuring the lighting, and realistically animating all objects and their interactions over time. Realism on a large scale is primarily a content creation problem. While all computer vision researchers have access to open-source engines that incorporate the latest advances in real-time rendering [50], there are no open-source virtual worlds with content that approaches the scale and realism of commercial productions.

Our approach is inspired by recent research that has demonstrated that ground-truth data for semantic segmentation can be produced for images from computer games without direct access to their source code or content [52]. The data collection techniques developed in this prior work are not sufficient for our purposes: they cannot generate data at video rate, do not produce instance-level segmentations, and do not produce dense correspondences, among other limitations. To create the presented benchmark, we have developed a new methodology for collecting ground-truth data from computer games, described in the next section.

## 3. Data Collection Methodology

To create the presented benchmark, we use Grand Theft Auto V, a modern game that simulates a functioning city and its surroundings in a photorealistic three-dimensional world. (The publisher of Grand Theft Auto V allows non-commercial use of footage from the game as long as certain conditions are met, such as non-commercial use and not distributing spoilers [53, 54].) A known way to extract data from such simulations is to inject a middleware between the game and its underlying graphics library via detouring [28]. The middleware acts as the graphics library and receives all rendering commands from the game. Previous work has adapted graphics debugging software for this purpose [52]. Since such software was designed for different use-cases, previous methods were limited in the frequency and granularity of data that could be captured. To create the presented benchmark, we have developed dedicated middleware that captures only data that is relevant to ground-truth synthesis, enabling it to operate at video rate and collect much richer datasets.

This work builds on fairly detailed understanding of real-time rendering. We provide an introductory overview of real-time rendering pipelines in the supplement and refer the reader to a comprehensive reference for more details [2].

**Slicing shaders.** To capture resources at video rate, we augment the game's shaders with additional inputs, outputs, and instructions. This enables tagging each pixel in real time with resource IDs as well as depth and transparency values. In order to augment the shaders at runtime, we employ dynamic software updating [27]. Dynamic software updates are used for patching critical software systems without downtime. Instead of shutting down a system, updating it, and bringing it back up, inactive parts are identified at runtime and replaced by new versions, while also transitioning the program state. This can be extremely challenging for complex software systems. We leverage the fact that shaders are distributed as bytecode blocks in order to be executed on the GPU, which simplifies static analysis and modification in comparison to arbitrary compiled binaries.

In particular, we start by slicing pixel shaders [61]. That is, we identify unused slots for inputs and outputs and instructions that modify transparency values, and ignore the remaining parts of the program. Note that source code for the shaders is not available and we operate directly on the bytecode. We broadcast an identifier for rendering resources as well as the $z$-coordinate to all affected pixels. To capture alpha values, we copy instructions that write to the alpha channel of the original G-buffer, redirect them to our G-buffer, and insert them right after the original instructions. We materialize our modifications via bytecode rewriting, a technique used for runtime modification of Java programs [47].

**Identifying objects.** The resources used to render an object correlate with the object's semantic class. By tracking resources, rendered images can be segmented into temporally consistent patches that lie within object parts and are associated with semantic class labels [52]. This produces annotations for semantic segmentation, but does not segment individual object instances. To segment the scene at the level of instances, we track transformation matrices used to place objects and object parts in the world. Parts that make up the same object share transformation matrices. By clustering patches that share transformation matrices, individual object instances can be identified and segmented.

Identifying object instances in this way has the added benefit of resolving conflicting labels for rendering resources that are used across multiple semantic classes. (E.g., wheels that are used on both buses and trucks.) To determine the semantic label for an object instance, we aggregate the semantic classes of all patches that make up the instance and take the majority vote for the semantic class of the object.

**Capturing 3D scene layout.** We record the meshes used for rendering (specifically, vertex and index buffers as well as the input layout for vertex buffers) and transformation matrices for each mesh. This enables us to later recover the camera position from the matrices, thus obtaining ground truth for visual odometry. We further use the vertex buffers to compute 3D bounding boxes of objects and transform the bounding boxes to the camera's reference frame: this produces the ground-truth data for 3D scene layout.

**Tracking objects.** Tracking object instances and computing optical flow requires associating meshes across successive frames. This poses a set of challenges:

1. Meshes can appear and disappear.
2. Multiple meshes can have the same segment ID.
3. The camera and the objects are moving.
4. Due to camera motion, meshes may be replaced by versions at different levels of detail, which can change the mesh, segment ID, and position associated with the same object in consecutive frames [37].

We tackle the first challenge by also recording the rendering resources used for meshes that either failed the depth test or lie outside the view frustum. We address the remaining challenges by formulating the association of meshes across frames as a weighted matching problem [20].

Let $V_f, V_g$ be nodes representing meshes rendered in two consecutive frames $f$ and $g$, and let $G = (V_f \cup V_g, E)$ be a graph with edges $E = \{e_{i,j} = (v_i \in V_f, v_j \in V_g)\}$, connecting every mesh in one frame to all meshes in the other. Let $s : V \to \mathcal{S}$ be the mapping from a mesh to its segment ID, $\mathcal{C}$ the set of semantic class labels, $\mathcal{D} \subseteq \mathcal{C}$ the set of semantic class labels that represent dynamic objects, and $c : V \to \mathcal{C}$ the mapping from meshes to semantic classes. Let $p : V \to \mathbb{R}^3$ be the mapping from a mesh to its position in the game world, given by the world matrix $W$ that transforms the mesh. Let $d(v_i, v_j) = \|p(v_i) - p(v_j)\|$ be the distance between positions of two meshes in the game world, and let $w : \mathcal{C} \to \mathbb{R}_0^+$ be a function for the class-dependent maximum speed of instances. Let $E_c = \{e_{i,j} : c(v_i) = c(v_j)\}$ and $E_k = \{e_{i,j} : s(v_i) = s(v_j)\}$ be the sets of edges associating meshes that share the same semantic class or segment ID, respectively. We define the sets $E_d = \{e_{i,j} : c(v_i) \in \mathcal{D}\} \cap E_c \cap E_k$ and $E_s = \{e_{i,j} : c(v_i) \in \mathcal{C} \setminus \mathcal{D}\} \cap E_c$ of edges associating meshes of dynamic objects and static objects, respectively. Finally, we define a set

$$E_m = \{e_{i,j} : d(v_i, v_j) < w(c(v_i))\} \cap (E_d \cup E_s) \quad (1)$$

and define a weight function on the graph:

$$w(v_i, v_j) = \begin{cases} d(v_i, v_j) & e_{i,j} \in E_m \\ 0 & \text{otherwise} \end{cases} \quad (2)$$

This leverages the previously obtained mappings of segments to semantic classes. Intuitively, we associate meshes by minimizing their motion between frames and prune associations of mismatched classes and mismatched segment IDs in the case of dynamic objects. Additionally, we cap motions depending on the mapped semantic classes (first term of Eq. 1). By solving the maximum weight matching problem on the graph, we associate meshes across pairs of consecutive frames. For keeping track of objects that are invisible for multiple frames, we extrapolate their last recorded motion linearly for a fixed number of frames and add their meshes to $V_f$ and $V_g$.

**Dense correspondences.** So far we have established correspondences between whole meshes, allowing us to track objects across frames. We now extend this to dense correspondences over the surfaces of rigid objects. To this end, we need to acquire dense 3D coordinates in the object's local coordinate system, trace transformations applied by shaders along the rendering pipeline in different frames, and invert these transformations to associate pixel coordinates in different frames with stable 3D coordinates in the object's original coordinate frame. To simplify the inversion, we disable tessellation shaders.

Let $\mathbf{x} \in \mathbb{R}^4$ be a surface point in object space, represented in homogenous coordinates. The transformation pipeline maps $\mathbf{x}$ to a camera-space point $\mathbf{s}$ via a sequence of linear transformations: $\mathbf{s} = CPVW\mathbf{x}$. For fast capture, we only record the world, view, and projection matrices $W, V, P$ for each mesh and the $z$-component (depth) of $\mathbf{s}$ for each pixel, as the clipping matrix $C$ and the $x, y$-components of $\mathbf{s}$ can be derived from the image resolution. By setting the $w$-component of $\mathbf{s}$ to 1 and inverting the matrices, we recover $\mathbf{x}$ for each pixel. To compute dense correspondences across frames $f$ and $g$, we obtain object-space points $\mathbf{x}$ from image-space points in $f$, and then use the transformation matrices of $g$ (obtained by associating meshes across $f$ and $g$) to obtain corresponding camera-space points in $g$.

**Inverting shader slices.** We now tackle dense correspondences over the surfaces of nonrigid objects. Nonrigid transformations are applied by vertex shaders. Unlike rigid correspondences, these transformations are not communicated to the graphics library in the form of matrices represented in standard format, and can be quite complex. Consider the set of vertices in the mesh of a person, represented in its local coordinate frame in a neutral pose. To transform the person to sit on a bench, the vertices are transformed individually by vertex shaders. The rasterizer interpolates the transformed mesh to produce 3D coordinates for every pixel depicting the person. To invert these black-box transformations, we have initially considered recording all input and output buffers in the pipeline and learning a non-parametric mapping from output to input via random forests [11]. This would have yielded approximate model-space coordinates for each pixel, rather than the precise, subpixel-accurate correspondences we seek.

We have instead developed an exact solution based on selectively executing the shader slices offline. The key idea is to use slices of a vertex shader for transforming meshes and inverting the rasterizer stage to map pixels back to their corresponding mesh location. We analyze the data flow within a vertex shader towards the output register of the transformed 3D points. Since we are only interested in the 3D coordinates, we can ignore (and remove) the computation of texture coordinates and other attributes, producing a slice of the vertex shader, which applies a potentially non-rigid transform to a mesh. Each transform, however, is rigid at the level of triangles. Furthermore, the order of vertices is preserved in the output of a shader slice, making the mapping from output vertices to input vertices trivial. The remaining step is to map pixels back to the transformed triangles, which reduces to inverting the linear interpolation of the rasterizer. Combining these steps, we map dense 3D points in camera coordinates to the object's local coordinate frame. As in the case of rigid transformations, this inversion is the crucial step. By chaining inverse transformations from one frame and forward transformations from another, we can establish dense subpixel-accurate correspondences over nonrigidly moving objects across arbitrary baselines.

## 4. Dataset

We have used the approach described in Section 3 to collect video sequences with a total of 254,064 frames at $1920 \times 1080$ resolution. The sequences were captured in five different ambient conditions: day (overcast), sunset, rain, snow, and night. All sequences are annotated with the following types of ground truth, for every video frame: pixelwise semantic categories (semantic segmentation), dense pixelwise semantic instance segmentations, instance-level semantic boundaries, object detection and tracking (the semantic instance IDs are consistent over time), 3D scene layout (each instance is accompanied by an oriented and localized 3D bounding box, also consistent over time), dense subpixel-accurate optical flow, and ego-motion (visual odometry). In addition, the metadata associated with each frame enables creating ground truth for additional tasks, such as subpixel-accurate correspondences across wide baselines (non-consecutive frames) and relative pose estimation for widely separated images.

The dataset is split into training, validation, and test sets, containing 134K, 50K, and 70K frames, respectively. The split was performed such that the sets cover geographically distinct areas (no geographic overlap between train/val/test) and such that each set contains a roughly balanced distribution of data acquired in different conditions (day/night/etc.) and different types of scenes (suburban/downtown/etc.). To perform the split, we clustered the recorded frames by geographic position and manually assigned clusters to the three sets. The dataset and associated benchmark suite are referred to as the VIsual PERception benchmark (VIPER).

**Statistical analysis.** We compare the statistics of VIPER to three benchmark datasets: Cityscapes [10], KITTI [19], and MS COCO [35]. The results of multiple analyses are summarized in Figure 2. First we evaluate the realism of the simulated world by analyzing the distributions of the number of categories and the number of instances present in each image in the dataset. For these statistics, our reference is the Cityscapes dataset, since the annotations in Cityscapes are the most accurate and comprehensive. As shown in Figure 2(a,b), the distribution of the number of categories per image in VIPER is almost identical to the distribution in Cityscapes. The Jensen-Shannon divergence (JSD) between these two distributions is 0.003: two orders of magnitude tighter than $\mathrm{JSD}\,(\mathrm{Cityscapes} \parallel \mathrm{COCO}) = 0.67$ and $\mathrm{JSD}\,(\mathrm{Cityscapes} \parallel \mathrm{KITTI}) = 0.69$. The distribution of the number of instances per image in VIPER also matches the Cityscapes distribution more closely by an order of magnitude than the other datasets: $\mathrm{JSD}\,(\mathrm{Cityscapes} \parallel \cdot) = (0.02; 0.19; 0.30)$ for $\cdot = (\mathrm{VIPER}; \mathrm{COCO}; \mathrm{KITTI})$. Further details are provided in the supplement.

Next we analyze the number of instances per semantic class. As shown in Figure 2(c), the number of semantic categories with instance-level labels in our dataset is 11, compared to 10 in Cityscapes and 7 in KITTI, while the number of instances labeled with pixel-level segmentation masks for each class is more than an order of magnitude higher.

Finally, we evaluate the realism of 3D scene layouts in our dataset. Figure 2(d) reports the distribution of vehicles as a function of distance from the camera in the three datasets for which this information could be obtained. The Cityscapes dataset again serves as our reference due to its comprehensive nature (data from 50 cities). The distance distribution in VIPER closely matches that of Cityscapes: $\mathrm{JSD}\,(\mathrm{Cityscapes} \parallel \mathrm{VIPER}) = 0.02$, compared to $\mathrm{JSD}\,(\mathrm{Cityscapes} \parallel \mathrm{KITTI}) = 0.12$.

**Perceptual experiment.** To assess the realism of VIPER in comparison to other synthetic datasets, we conduct a perceptual experiment. We sampled 500 random images each from VIPER, SYNTHIA [55], Virtual KITTI [17], the Freiburg driving sequence [41], and the Urban Canyon dataset (regular perspective camera) [64], as well as Cityscapes as a real-world reference. Pairs of images from separate datasets were selected at random and shown to Amazon Mechanical Turk (MTurk) workers who were asked to pick the more realistic image in each pair. Each MTurk job involved a batch of ~100 pairwise comparisons, balanced across conditions and randomized, along with sentinel pairs that test whether the worker is attentive and diligent. Each job is performed by 10 different workers, and jobs in which any sentinel pair is ranked incorrectly are pruned. Each pair is shown for a timespan chosen at random from $\{\frac{1}{8}, \frac{1}{4}, \frac{1}{2}, 1, 2, 4, 8\}$ seconds. For each timespan and pair of datasets, at least 50 distinct image pairs were rated. This experimental protocol was adopted from concurrent work on photographic image synthesis [9].

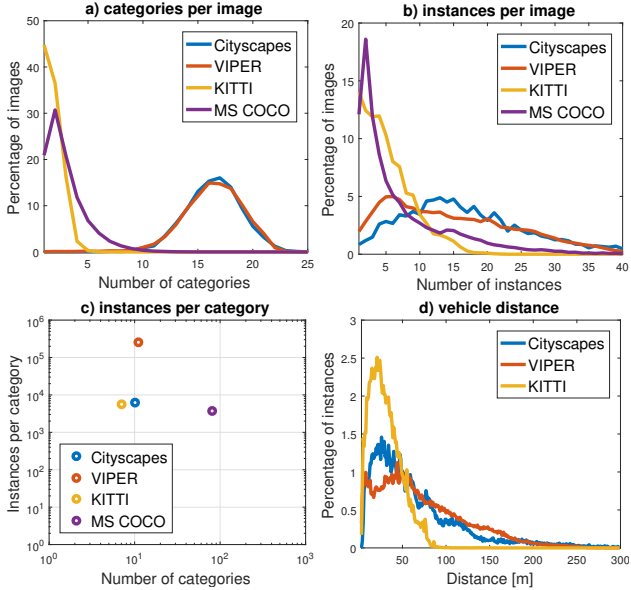

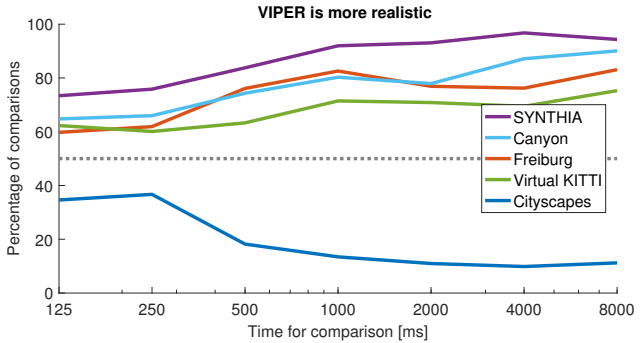

Figure 2. **Statistical analysis** of the VIPER dataset in comparison to Cityscapes, KITTI, and MS COCO. (a,b) Distributions of the number of categories and the number of instances present in each image. (c) Number of instances per semantic class. (d) Distributions of vehicles as a function of distance from the camera.

The results are shown in Figure 3. VIPER images were rated more realistic than all other synthetic datasets. The difference is already apparent at 125 milliseconds, when the VIPER images are rated more realistic than other synthetic datasets in 60% (vs Freiburg) to 73% (vs SYNTHIA) of the comparisons. This is comparable to the relative realism rate of real Cityscapes images vs VIPER at this time (65%). At 8 seconds, VIPER images are rated more realistic than all other synthetic datasets in 75% (vs Virtual KITTI) to 94% (vs SYNTHIA) of the comparisons. Surprisingly, the votes for real Cityscapes images versus VIPER at 8 seconds are only at 89%, lower than the rates for VIPER versus two of the other four synthetic datasets.

## 5. Baselines and Analysis

We have set up and analyzed the performance of representative methods for semantic segmentation, semantic instance segmentation, visual odometry, and optical flow. Our primary goals in this process were to further validate the realism of the VIPER benchmark, to assess its difficulty relative to other benchmarks, to provide reference baselines for a number of tasks, and to gain additional insight into the performance characteristics of state-of-the-art methods.

Figure 3. **Perceptual experiment.** Results of pairwise A/B tests with MTurk workers, who were asked to select the more realistic image in a pair of images, given different timespans. The dashed line represents chance. VIPER images were rated more realistic than images from other synthetic datasets.

**Semantic segmentation.** Our first task is semantic segmentation and our primary measure is mean intersection over union (IoU), averaged over semantic classes [10, 16]. We have evaluated two semantic segmentation models. First, we benchmarked a fully-convolutional setup of the ResNet-50 network [26, 36]. Second, we benchmarked the Pyramid Scene Parsing Network (PSPNet), an advanced semantic segmentation system [65]. The results are summarized in Table 1.

| Method | day | sunset | rain | snow | night | all |
|---|---|---|---|---|---|---|
| FCN-ResNet | 57.3 | 53.5 | 52.4 | 54.3 | 57.8 | 55.1 |
| PSPNet [65] | 73.6 | 68.5 | 65.6 | 66.4 | 70.3 | 68.9 |

Table 1. **Semantic segmentation.** Mean IoU in each environmental condition, as well as over the whole test set.

We draw several conclusions. First, the relative performance of the two baselines on VIPER is consistent with their relative performance on the Cityscapes validation set (PSPNet is ahead by ∼14 points on both datasets). Second, VIPER is more challenging than Cityscapes: while PSPNet is above 80% on Cityscapes, it's at 69% on VIPER. We view this additional headroom as a benefit to the community, given that the performance of the leading methods on Cityscapes rose by more than 10 points in the last year. Third, the relative performance in different conditions is broadly consistent across the two methods: for example, day is easier than sunset, which is easier than rain. Fourth, the methods do not fail in any condition, indicating that contemporary semantic segmentation systems, which are based on convolutional networks, are quite robust when diverse training data is available.

**Semantic instance segmentation.** To measure semantic instance segmentation accuracy, we compute region-level average precision (AP) for each class, and average across classes and across 10 overlap thresholds between 0.5 and 0.95 [10, 35]. We benchmark the performance of Multi-task Network Cascades (MNC) [12] and of Boundary-aware Instance Segmentation (BAIS) [25]. Each system was trained in two conditions: (a) on the complete training set and (b) on day and sunset images only. The results are shown in Table 2.

| Method | Train condition | day | sunset | rain | snow | night | all |
|---|---|---|---|---|---|---|---|
| MNC [12] | day & sunset | 9.0 | 7.6 | 6.7 | 5.2 | 2.3 | 6.2 |
| MNC [12] | all | 9.1 | 7.7 | 9.6 | 6.8 | 5.4 | 7.7 |
| BAIS [25] | day & sunset | 13.3 | 12.4 | 11.2 | 8.2 | 3.1 | 9.6 |
| BAIS [25] | all | 14.7 | 11.5 | 14.2 | 10.8 | 11.6 | 12.6 |

Table 2. **Semantic instance segmentation.** We report AP in each environmental condition as well as over the whole test set. Models trained only on day and sunset images do not generalize well to other conditions.

We first observe that the relative performance of MNC and BAIS on VIPER is consistent with their relative performance on Cityscapes, and that VIPER is more challenging than Cityscapes in this task as well (12.6 AP for BAIS on VIPER vs 17.4 on Cityscapes). Furthermore, we see that the performance of systems that are only trained on day and sunset images drops in other conditions. The performance drop is present in all conditions and is particularly dramatic at night. Note that we matched the number of training iterations in the 'day & sunset' and 'all' regimes, so the 'day & sunset' models are trained for a proportionately larger number of epochs to compensate for the smaller number of images.

A performance analysis of instance segmentation accuracy as a function of objects' distance from the camera is provided in Figure 4.

**Optical flow.** To evaluate the accuracy of optical flow algorithms, we use a new robust measure, the Weighted Area Under the Curve. The measure evaluates the inlier rates for a range of thresholds, from 0 to 5 px, and integrates these rates, giving higher weight to lower-threshold rates. The thresholds and their weights are inversely proportional. The precise definition is provided in the supplement. This measure is a continuous analogue of the measure used for the KITTI optical flow leaderboard [19]. Instead of selecting a specific threshold (3 px), we integrate over all thresholds between 0 and 5 px and reward more accurate estimates within this range.

We benchmark the performance of four well-known optical flow algorithms: LDOF [7], EpicFlow [51], FlowFields [3], and FlowNet [14]. For all methods, we used the publicly available implementations with default parameters.

The results are summarized in Table 3. The relative ranking of the four methods on VIPER is consistent with their relative accuracy on the KITTI optical flow dataset. The poor performance of FlowNet is likely due to its exclusive training on a different dataset; we expect that training this model (or its successor [29]) on VIPER will yield much better results. Overall the results indicate that VIPER is more challenging than KITTI, even in the daytime condition. We attribute this to the more varied and complex nature of our scenes (see Figure 2), and the density and precision of our ground truth (including on nonrigidly moving objects, on thin structures, around boundaries, etc.). For all methods, accuracy degrades markedly in the rain, in the presence of snow, and at night.

| Method | day | sunset | rain | snow | night | all |
|---|---|---|---|---|---|---|
| FlowNet [14] | 41.3 | 41.9 | 28.2 | 40.4 | 33.7 | 37.1 |
| LDOF [7] | 69.8 | 60.3 | 44.4 | 57.0 | 53.3 | 56.9 |
| EpicFlow [51] | 76.8 | 67.1 | 52.4 | 65.3 | 59.7 | 64.2 |
| FlowFields [3] | 78.4 | 68.1 | 52.1 | 66.3 | 60.2 | 65.0 |

Table 3. **Optical flow.** We report the weighted area under the curve, a robust measure that is analogous to the inlier rate at a given threshold but integrates over a range of thresholds (0 to 5 px) and assigns higher weight to lower thresholds. Higher is better.

We further analyze the performance of EpicFlow, which is commonly used as a building block in other optical flow pipelines (e.g., FlowFields). Specifically, we investigate the accuracy of optical flow estimation as a function of object type, object size, and displacement magnitude. The results of this analysis are shown in Figure 5. We see that the most significant challenges are posed by very large displacements, very large objects (primarily people and vehicles that are close to the camera), ground and vehicle motion, and adverse weather.

**Visual odometry.** For evaluating the accuracy of visual odometry algorithms, we follow Geiger et al. [19] and measure the rotation errors of sequences of different lengths and speeds. We benchmark two state-of-the-art systems that represent different approaches to monocular visual odometry: ORB-SLAM2, which tracks sparse features [45], and DSO, which optimizes a photometric objective defined over image intensities [15]. For fair comparison we run both methods without loop-closure detection. We tuned hyperparameters for both methods on the training set. To account for non-deterministic behavior due to threading, we ran all configurations 10 times on all test sequences.

The results are summarized in Table 4. The performance of the tested methods is highest in the easy *day* setting and decreases in adverse conditions. We identified a number of phenomena that affect performance. For example, detecting keypoints is harder with less contrast (*snow*), in areas lit by moving headlights (*night*), or in the presence of lens flares (*sunset*). Reflections on puddles (*rain*) often yield large sets of keypoints that are consistent with incorrect pose hypotheses. In all conditions, the accuracy is much lower than corresponding results on KITTI, indicating that the VIPER visual odometry benchmark is far more challenging. We attribute this to the different composition of VIPER scenes, which include wider streets with fewer features at close range, more variation in camera speed, and a higher concentration of dynamic objects (see Figure 2).

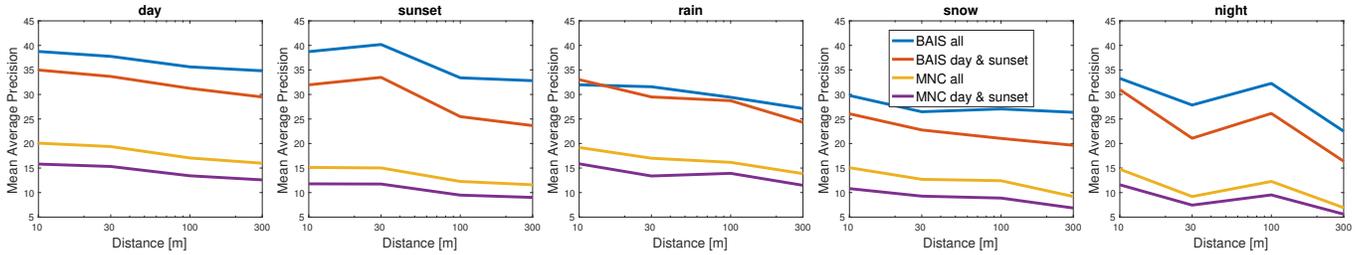

Figure 4. **Analysis of instance segmentation performance** as a function of the objects' distance from the camera, in different conditions. All methods perform best on objects within 10 meters. Accuracy deteriorates with distance.

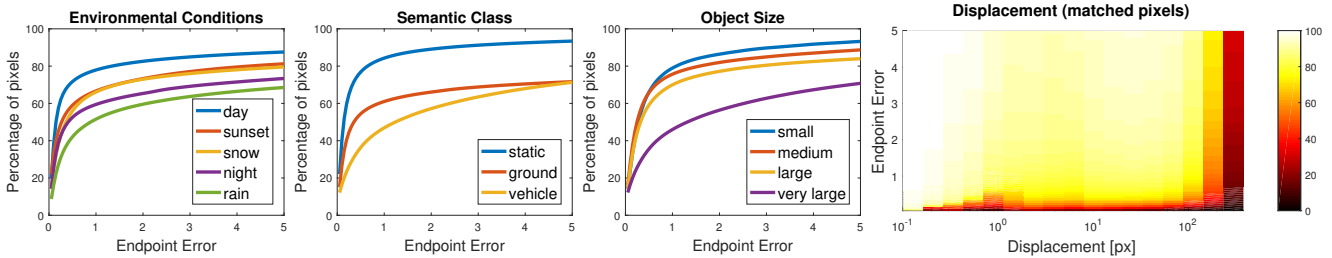

Figure 5. **Analysis of optical flow performance**, conducted on EpicFlow. Left to right: effect of environmental condition, object type, object size, and displacement.

| Method | day | sunset | rain | snow | night | all |
|---|---|---|---|---|---|---|
| ORB-SLAM2 | 0.253 | 0.420 | 0.294 | 0.381 | 0.356 | 0.341 |
| DSO | 0.204 | 0.270 | 0.244 | 0.259 | 0.250 | 0.245 |

Table 4. **Visual odometry.** We report rotation errors (in deg/m) for ORB-SLAM2 [45] and for DSO [15].

An interesting opportunity for future work is to integrate visual odometry with semantic analysis of the scene, which can help prune destabilizing keypoints on dynamic objects and can restrain scale drift by estimating the scale of objects in the scene [32]. The presented benchmark provides integrated ground-truth data that can facilitate the development of such techniques.

## 6. Conclusion

We have presented a new benchmark suite for visual perception. The benchmark is enabled by ground-truth data for both low-level and high-level vision tasks, collected for more than 250 thousand video frames in different environmental conditions. We hope that the presented benchmark will support the development of techniques that leverage the temporal structure of visual data and the complementary nature of different visual perception tasks. We hope that the availability of ground-truth data for all tasks on the same video sequences will support the development of robust broad-competence visual perception systems that construct and maintain effective models of their environments.

The dataset will be released upon publication. Ground-truth data for the test set will be withheld and will be used to set up a public evaluation server and leaderboard. In addition to the baselines presented in the paper, we plan to provide reference baselines for additional tasks, such as 3D layout estimation, and to set up challenges that evaluate performance on integrated problems, such as temporally consistent semantic instance segmentation, tracking, and 3D layout estimation in video.

## Acknowledgements

We thank Qifeng Chen for the MTurk experiment code, Fisher Yu for the FCN-ResNet baseline, the authors of PSP-Net for evaluating their system on our data, and Jia Xu and René Ranftl for help with optical flow and visual odometry analysis. SRR was supported in part by the European Research Council under the European Union's Seventh Framework Programme (FP/2007-2013) / ERC Grant Agreement No. 307942.